\documentclass{article}

\usepackage{arxiv}

\usepackage[utf8]{inputenc} 
\usepackage[T1]{fontenc}    
\usepackage{hyperref}       
\usepackage{url}            
\usepackage{booktabs}       
\usepackage{amsfonts}       
\usepackage{nicefrac}       
\usepackage{microtype}      
\usepackage{lipsum}
\usepackage{graphicx}
\graphicspath{ {./images/} }

\usepackage{amsmath}
\usepackage{amssymb}
\usepackage{mathtools}
\usepackage{amsthm}
\usepackage{thmtools}

\usepackage[capitalize,noabbrev]{cleveref}
\usepackage{nicematrix}

\usepackage{booktabs}
\usepackage{colortbl}
\definecolor{best}{RGB}{175,216,230} 
\definecolor{secondbest}{RGB}{220,220,220} 

\theoremstyle{plain}
\newtheorem{theorem}{Theorem}[section]
\newtheorem{proposition}[theorem]{Proposition}

\newtheorem{assumption}[theorem]{Assumption}
\theoremstyle{remark}
\newtheorem{remark}[theorem]{Remark}

\usepackage[textsize=tiny]{todonotes}
\usepackage{wrapfig}

\runningTitle{Causal Cellular Context Transfer Learning}
\title{Causal Cellular Context Transfer Learning (C$^3$TL): \\
\large An Efficient Architecture for Prediction of Unseen Perturbation Effects}

\author{
 Michael Scholkemper \\
  Statistics and Machine Learning,\\
  German Center for Neurodegenerative Diseases  (DZNE), Bonn, Germany.\\
  \texttt{michael.scholkemper@dzne.de} 
   \And
 Sach Mukherjee \\
  Statistics and Machine Learning,
  German Center for Neurodegenerative Diseases (DZNE) Bonn, Germany,\\
  University of Bonn, Germany \& University of Cambridge, United Kingdom.
}

\begin{document}
\maketitle
\begin{abstract}
Predicting the effects of chemical and genetic perturbations on quantitative cell states is a central challenge in computational biology, molecular medicine and drug discovery. 
Here we propose an efficient neural framework for perturbation effect prediction that exploits the structured nature of biological interventions and specific inductive biases/invariances. Our approach leverages available information  concerning perturbation effects to
allow generalization to novel contexts and requires only widely-available bulk molecular data.
Extensive testing, comparing
 predictions of context-specific perturbation effects  against
real, large-scale interventional experiments, 
demonstrates 
accurate prediction in new contexts.
The proposed approach is 
data-, parameter- and compute-efficient yet 
competitive with state-of-the-art, large-scale foundation models.
Focusing on robust bulk signals and efficient  architectures, we show that accurate prediction of perturbation effects is possible 
by sharing information across contexts, hence opening up ways to leverage causal learning approaches in biomedicine generally. 
\end{abstract}

\section{Introduction}
\label{sec:intro}

The ability to predict how cells respond to external perturbations, such as genetic modifications, treatment with compounds, or environmental changes, is central to contemporary biomedical research. From a machine learning (ML) point of view, modelling effects under perturbation is a fundamentally causal question, concerning change under intervention and potentially large distribution shifts brought about by perturbation. 
Due to the fact that biological and clinical phenotypes are driven by the underlying cellular state, understanding changes under perturbation is 
essential across biology and medicine. As the volume of biological data grows, AI and ML approaches are emerging as key elements in 
workflows aimed at understanding complex cellular states and their modulation.
AI- and ML-based \textit{in silico} models in particular have the potential to 
allow virtual analogues to real experiments at vastly lower cost and larger scale and coverage.

In recent years, experimental	approaches have emerged  that allow causal effects on gene expression to be studied at genome-wide scale
\cite{dixit2016perturb,replogle2022mapping, nadig2025transcriptome}. However, due to a range of  complex biophysical and biochemical mechanisms, changes under perturbation are dependent on cellular, genetic and epigenetic context. This means that the same perturbation -- say knockout of a specific gene, or treatment with a specific chemical compound -- may induce quite different changes in cells of type $A$ or $B$, or in different genetic backgrounds. 
This in turn limits brute-force experimental approaches, since it remains infeasible to cover all possible interventions on all possible cell types/contexts.

Against this background, significant progress has been made in recent years using deep generative models
to analyze and integrate perturbation data. 
A prominent example is the Compositional Perturbation Autoencoder (CPA) \cite{lotfollahi2023predicting}, which learns to disentangle perturbation effects from cell types in single-cell RNA sequencing (scRNA-seq) data, enabling the prediction of cellular responses to unseen combinations. More recently, the field has seen a shift towards massive ``foundation models'', exemplified by the \emph{State} model 
\cite{adduri2025predicting}. These large-scale  architectures, typically based on transformers and related long-context models, 
are trained on vast atlases of genetic and single-cell data to generalize across diverse biological contexts.

However, the reliance on single-cell data and massive model architectures presents substantial barriers. While single-cell resolution offers detailed insights into cellular heterogeneity, it suffers from high technical noise, sparsity, and significant costs compared to traditional RNA sequencing (so-called ``bulk'' measurements in which data are acquired from many cells together).
Bulk expression profiles, through averaging over cell populations, provide a robust and cost-effective signal that remains the standard for many clinical and pharmaceutical applications.

The heterogeneity of biological and medical problems means that data collected in a specific context (specific cells, disease state etc.) are highly informative. The scope of truly zero-shot approaches remains unclear and there remains a need for model adaptation and re-training including integration into domain-specific workflows. 
However, with very large models such adaptation and/or integration can be burdensome.
Computational issues may be critical 
in the context of  the resource constraints typical of academic laboratories and hospitals. Training or even fine-tuning very large models 
(e.g. with ${\sim}10^9$ parameters or more)
may require clusters of high-end GPUs, creating a disparity where advanced  tools are accessible only to a handful of research centers. In contrast, efficient, targeted models that can run on widely-available hardware offer a path to make these technologies 
accessible for a wider community
and in ways that can open the way to richer integration and
adaptation within scientific models and workflows in the future.

In this work, we address these challenges by proposing a method 
by which changes under perturbations in a novel context are modelled by transferring information from other contexts. 
Our approach requires only commonly available bulk gene expression data.
 We argue that for many downstream tasks, the robustness of bulk data combined with efficient neural models can achieve competitive performance without the overhead of single-cell data or 
 truly large-scale
 foundation models. Our contribution highlights the advantage of ``right-sized'' AI modelling in biomedicine, 
 and supports the notion that in some cases  powerful generalization capabilities can be achieved using (relatively) limited data and with (relatively) moderate 
 computational effort. 

\section{Related Work}

\textbf{Causal structure learning and causal effect prediction.}
A rich literature in ML and statistics focuses on learning casual structures from data \cite{heinze2018causal}, including 
manifold learning perspectives \cite{hill2019causal}
and recent work leveraging a range of neural architectures \cite{lopez2022large,lippe2022,ke2022learning,lagemann2023deep}.
Many of these approaches exploit interventional information and can be viewed as generalizing to novel interventional regimes, by providing information on effects downstream of specific nodes. The role of context in this literature is typically implicit and appears via the data that is used to train the models. Causal effect prediction \cite{pfister2022identifiability} concerns regression-type problems rather than graph learning, but also seeks to generalize to novel interventions on a given system.

\textbf{Generative Modeling of Cellular Perturbations.}
In the context of large-scale systems biology, deep generative frameworks, going beyond linear and generalized linear models,
have attracted much recent attention. The seminal work of scGen \cite{lotfollahi2019scgen} introduced the concept of latent space arithmetic for perturbation prediction, demonstrating that deep autoencoders could learn to shift cell states along a ``perturbation vector''. This paradigm was significantly expanded by the Compositional Perturbation Autoencoder (CPA) \cite{lotfollahi2023predicting}, which utilized adversarial alignment to disentangle cell-type identity from perturbation effects, enabling the prediction of unseen drug combinations. Further extensions such as ChemCPA \cite{hetzel2022predicting} integrated chemical structural information to generalize to unseen compounds. Apart from autoencoder-based methods, graph neural networks have also been employed; for instance, GEARS \cite{roohani2023gears} leverages gene interaction graphs to predict the outcomes of combinatorial genetic perturbations.

\textbf{Foundation Models and Large-Scale Learning.}
Concomitant with the rise of transformers in natural language processing, the field has moved towards large foundation models. Prominent examples include scGPT \cite{cui2024scgpt} and the \emph{State} model \cite{adduri2025predicting}. These models
(based on transformers and related architectures) are trained on comprehensive single-cell atlases to predict responses across diverse biological contexts. 
These models are an important area of contemporary research, but
their development relies  on  large ($>10^9$ parameters) architectures and extensive GPU clusters.

\textbf{Bulk Gene Expression and Transfer Learning.}
Despite the dominance of single-cell methods in recent literature, bulk gene expression profiling remains the industry standard for high-throughput screening, epitomized by the massive L1000 dataset \cite{subramanian2017l1000}. Prior computational efforts for bulk data often utilized matrix and tensor factorization to impute missing experiments in the drug-cell-gene data cube \cite{hodos2018cell,iwata2019predicting}. However, these methods typically lack the capacity to explicitly disentangle biological state from perturbation effects in a transfer learning framework.

\textbf{Novel contributions.}
Our work makes a number of novel contributions relative to the lines of work outlined above:
\begin{itemize}
\item {\it Contextual generalization.} In contrast to many existing approaches in causal structure learning and causal effect prediction, we focus on the setting in which all interventions of interest have already been seen in some contexts, with the question of interest being generalization to novel {\it contexts}. This focus is motivated by the biological application domain, where emerging experimental protocols are allowing empirical work spanning large numbers of perturbations \cite{replogle2022mapping, nadig2025transcriptome}, but where the role of context is paramount. 

\item {\it Quantitative effect prediction.} In contrast to work on causal structure learning, we focus on the prediction of quantitative effects. This is motivated by downstream tasks concerning prediction of phenotypes and AI-in-the-loop approaches. 

\item {\it Compute-, parameter- and data-efficiency.} In contrast to recent foundation models (FMs) we focus on relatively efficient architectures that are arguably particuarly amenable to academic and biomedical use.
In empirical work, we study also performance in few-shot, limited data settings -- including the case where very little information is available on a specific target context -- 
that better reflect real-world use cases.

\end{itemize}

\section{Methods}

In this Section, we develop key ideas in the proposed methodology at a conceptual level, deferring details concerning architectures, training and inference to the sequel. 
We first develop the main ideas assuming complete information.
Prediction of change under intervention in a novel context with potentially influential hidden factors is not possible without additional assumptions since -- absent further assumptions -- the causal effect could be arbitrarily different in the target context. We introduce assumptions under which we show 
existence of an encoder-decoder type architecture that can be trained on only available data and information. This allows us to practically address the setting in which we aim to model perturbation effects in real-world, incomplete information use-cases.

\subsection{Problem statement and notation}
We focus on the setting in which we are given (bulk) multivariate $d$-dimensional observations 
$(x_p^c)_{p \in \mathcal{P}, c \in \mathcal{C}}, \, x_p^c \in \mathbb{R}^d$ under different perturbations $p \in \mathcal{P}$ and in different cellular contexts $c \in \mathcal{C}$. Our aim is to predict the effect of the {\it same} perturbations 
$p \in \mathcal{P}$
in a {\it novel} context $c' \notin \mathcal{C}$ (i.e. where we have seen the effect of the perturbations only in other contexts/cells).
This is motivated by current biological and medical use-cases, where a dictionary of perturbations (e.g. genetic or compound perturbation libraries) can be defined and tested in a limited fashion across a few contexts, 
but where it remains infeasible 
to cover every possible context (cell type/genetic/epigenetic background).
We assume that there are finitely many possible perturbations and also finitely many possible contexts. 

Our specific quantitative target will be the difference between expression under
the perturbation with respect to a base, unperturbed state.
This requires a type of causal generalization that involves quantities that cannot be directly observed (and that may not be identifiable); to aid exposition, we denote theoretical quantities  (that are needed to understand the system but that will not be estimated in practice) with a star and below we will aim to arrive, step-by-step, at practically-applicable formulations in which such ``starred'' quantities do not appear.

\subsection{Complete Data}
For ease of exposition, we start by assuming we have complete data and access to various hidden (``starred'') quantities. 
We will then show in the sequel how (under additional assumptions) we can arrive at a formulation
that gives the required outputs using only available inputs. 

\bigskip

{\bf Cell-level model.}

\begin{assumption}
    Let $X_p^c {\in} \mathbb{R}^d$ be the random variable (RV) describing the $d$-dimensional readouts 
    of the cells in context $0 \leq c \leq C$ under perturbation $0 \leq p \leq P$. 
    We assume that $X_p^c$ is distributed as:
    $$
    X_p^c = \mathcal{T}(p,c, \mathcal{M}(c)) + \mathcal{M}(c) + \epsilon
    $$ 
    where $\mathcal{M}(c)$ is a RV describing biological base variation in the cell population, $\mathcal{T}(p,c, M(c))$ is an RV 
    describing the effect of perturbation $p$ 
    on a cell of type/context $c$ with a basal state of $M(c)$, and $\epsilon$ is zero-mean, context-independent  noise.
    Further, let $X_0^c$ be an unperturbed base state --- that is, $\mathcal{T}(0,c, \mathcal{M}(c)) = 0$.
\end{assumption}

Essentially, this assumes that the generative process leading to the readouts consists of sampling a cell basal state $\mathcal{M}(c)$ and subsequently sampling a perturbation $\mathcal{T}(p,c,\mathcal{M}(c))$ conditional on this basal state.
Note that in this formulation, the perturbation effect $\mathcal{T}$ depends on not only perturbation $p$ and context $c$, but also on the base expression $\mathcal{M}(c)$. This captures the biophysical fact that the effect of a perturbation depends on the actual, realized chemical state of the cell. 

\begin{figure*}[t]
\vskip 0.2in
\begin{center}
\resizebox{\textwidth}{!}{\includegraphics[width=\textwidth]{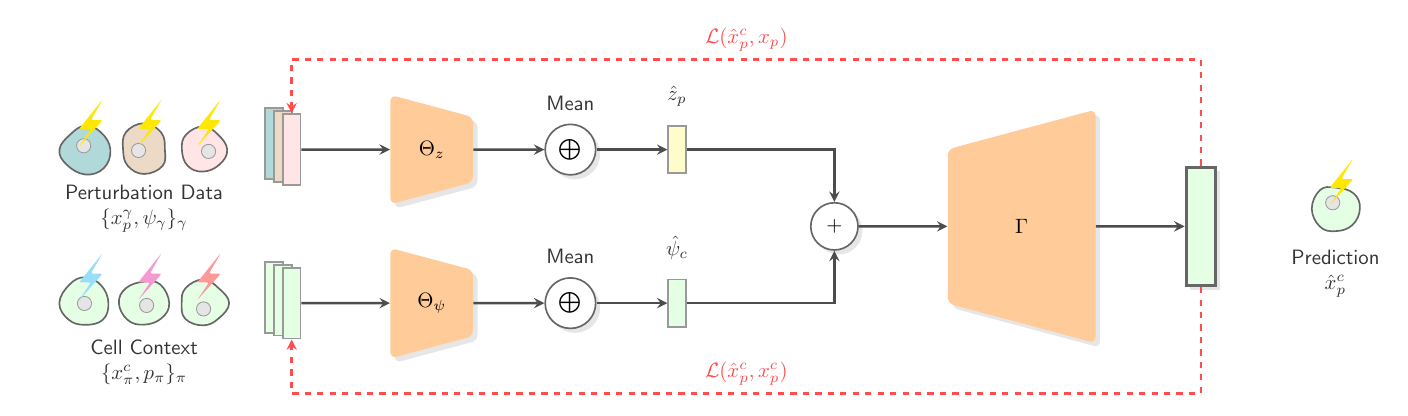}}
\caption{\textbf{Proposed architecture.} The model compresses the high-dimensional gene expression input $\{x^\gamma_p\}_\gamma$ from multiple contexts (indexed by $\gamma$) but under the {\it same} perturbation $p$ into a lower-dimensional latent representation $\hat{z}_p$ of the perturbation itself. Similarly, it forms a representation $\hat{\psi}_c$ of the context from all the different perturbations (indexed by $\pi$) observed in the specific context $c$. From these latent representations, it then reconstructs the original gene expression $\hat{x}^c_p$ using a decoder $\Gamma$.}
\label{fig:autoencoder}
\end{center}
\vskip -0.2in
\end{figure*}

\bigskip
{\bf Bulk perturbation effect.}
Now consider $\mathbb{E}[X_p^c] - \mathbb{E}[X_0^c]$, i.e.
the bulk-level difference of expression under perturbation $p$ from the base state.
This is the  $d$-dimensional perturbation effect that will form our prediction target.
To aid exposition, for the base state 
let the RV corresponding to $\mathcal{M}(c)$ be $\mathcal{M}'(c)$ (i.e.
$\mathcal{M}(c), \mathcal{M}'(c)$ are RVs with the same distribution)
and similarly let $\epsilon'$ be an RV with the same distribution as 
$\epsilon$.

We have that:
\begin{align}\label{eq:true_pert_effect_equals_diff_of_expectation}
\begin{split}
\mathbb{E}[X_p^c] - \mathbb{E}[X_0^c] &= \mathbb{E}[\mathcal{T}(p,c, \mathcal{M}(c)) + \mathcal{M}(c) + \epsilon] 
- \mathbb{E}[\mathcal{T}(0,c, \mathcal{M}'(c)) + \mathcal{M}'(c) + \epsilon']\\
&= \mathbb{E}[\mathcal{T}(p,c, \mathcal{M}(c))] - \mathbb{E}[\mathcal{T}(0,c, \mathcal{M}'(c))] 
+ \mathbb{E}[\mathcal{M}(c)] - \mathbb{E}[\mathcal{M}'(c)] + \mathbb{E}[\epsilon] - \mathbb{E}[\epsilon']\\
&= \mathbb{E}[\mathcal{T}(p,c, \mathcal{M}(c))] \\
& = \mathcal{T}^*(p,c)
\end{split}
\end{align}
We refer to $\mathcal{T}^*(p,c)$ as the {\it true perturbation effect} and, as seen above, this equals the difference of the expected expression under perturbation $p$ from the unperturbed base state. Notice that $\mathcal{T}^*$ is not a random quantity and that while $\mathcal{T}^*(\cdot, \cdot)$ may be a highly non-trivial function, it is nonetheless dependent on only two things: (i) a representation of the perturbation itself 
(not the perturbation effect in any specific context, but rather the perturbation itself in general) 
and (ii) a representation of the context.
In particular, it does not any more depend on the stochastic realization of the cell basal state $\mathcal{M}(c)$.

\bigskip
{\bf Towards a decoder architecture.}
Thus, the true perturbation effect $\mathcal{T}^*$
is a fixed, non-random quantity that depends only on the perturbation  $p$ and context  $c$.
This in turn means that given non-degenerate encodings $z_p^*, \psi_c^*$ of the perturbation and context respectively (i.e. such that $p \neq p' \iff z_p^* \neq z_{p'}^*$ and similarly for $\psi_c^*$),
there must exist a function $\mathcal{D}^*$ transforming the pair $(z_p^*,\psi_c^*)$ to the actual 
$d$-dimensional expression effect:
\begin{equation}
    \mathcal{D}^* : (z_p^*, \psi_c^*) \mapsto \mathcal{T}^*(p,c)
\end{equation}
This suggests the possibility of a decoder-type architecture that takes a latent representation of perturbation and
context to yield an estimate of the required perturbation effect. However, in practice, we do not have access to \textbf{all} perturbations and \textbf{all} cellular contexts.
Furthermore, under the foregoing  general formulation, absent additional assumptions, a novel context may be arbitrarily different from previously seen data (since so far there is no restriction on the term $\mathcal{T}^*(p,c)$ or
on $\mathcal{D}^*$).
Below, we introduce additional assumptions to allow training and inference in practice. 

\subsection{Generalizing to novel contexts}

Next, we introduce a key assumption 
concerning a type of causal manifold structure and show how this leads to a practical encoder-decoder 
formulation that can be trained on bulk data as would be available in practice.

\newpage
\begin{assumption}[Causal Manifold Assumption or CMA]\label{assumption:CMA}
    There exists an invertible function $f^* : \mathbb{R}^{q^*} \! \! \rightarrow \! \mathbb{R}^d$ that maps coordinates on a lower dimensional manifold to the ambient expression space in the following way:
    $$
     f^*(z_p^* + \psi_c^*) = \mathcal{T}^*(p,c), 
    $$
    where $z_p^* \in \mathbb{R}^{q^*}$ and $\psi_c^* \in \mathbb{R}^{q^*}$ are latent representations
    of perturbation and context respectively.
    \end{assumption}

CMA assumes there exists a latent causal manifold of dimension $q^*$
on which each perturbation $p$ has a general position $z_p^*$ which, for context $c$, is modified by a 
linear shift $\psi_c^*$. The position after the shift corresponds, in the ambient gene expression space, to the 
$d$-dimensional
context-specific effect.

The function $f^*$ may be a highly complex, non-linear function. Hence, 
recovering the true latent causal manifold in terms of $z^*_p$ and $\psi_c^*$ from data may be a challenging task. 
Furthermore, we do not actually have access to the true latent representations $z^*_p$ and $\psi_c^*$.

We next show that given suitable representations of perturbation and context, we can leverage CMA to define an end-to-end learning scheme that does {\it not}  require access to the true latent representations, and that is applicable in practice given limited bulk data.

\medskip
{\bf Perturbation and context representations.}
CMA (Assumption \ref{assumption:CMA}) is stated in terms of theoretical latent representations $z^*_p$ and $\psi_c^*$ that we will not have access to in practice. 
Suppose instead that we have access  to encodings $z_p \in \mathbb{R}^{q_z}$ and $\psi_c \in \mathbb{R}^{q_\psi}$.
These encodings could be very different from the corresponding latents 
$z^*_p, \psi_c^*$ but should be related to the true latents in the following way: 
\begin{equation}\label{eq:def_correction}
    z_p^* = h_z^*(z_p), \quad \psi_c^* = h_\psi^*(\psi_p).
\end{equation}
where $h_z^* : \mathbb{R}^{q_z} \rightarrow \mathbb{R}^{q^*}$ and $h_\psi^* : \mathbb{R}^{q_\psi} \rightarrow \mathbb{R}^{q^*}$ are functions mapping the available encodings to the true latents. 
These functions are starred because while we require their existence, we will not need to estimate them directly.
The assumption of {\it existence} of these functions is very mild: We can quickly verify that such embeddings must exist in our setting by taking $q_z = P$ and $z_p$ as a one-hot encoding of the perturbation and similarly $q_\psi = C$ and $\psi_c$ as a one-hot encoding of the context (recall that there are finitely many perturbations and contexts, hence this construction is always possible). 

We note that the functions 
$h_z^*, h_\psi^*$
may be arbitrarily complex and moreover supervised learning is not possible without access to the true latents.
However, we do not aim to actually learn 
$h_z^*, h_\psi^*$; rather, as we see next,
their existence implies a formulation
in which the available 
encodings can be used to directly predict the required perturbation effects.

\medskip
{\bf A practical encoder/decoder.}
Under the foregoing assumptions, we can adopt a learning scheme as follows. 
We can predict the perturbation effect 
as:
\begin{eqnarray*}
\mathcal{T}^*(p,c) & \overset{CMA}{=} & f^*(z_p^* + \psi_c^*) \\
& \overset{(\ref{eq:def_correction})}{=}&  f^*(h_z^*(z_p) + h_\psi^*(\psi_c)) \\
& = &  \mathcal{D}(z_p, \psi_c),
\end{eqnarray*}
where $\mathcal{D}  : \mathbb{R}^{q_z} {\times} \mathbb{R}^{q_\psi} {\rightarrow }\mathbb{R}^{d}$
is a decoder-type function.

A key observation is the decoder $\mathcal{D}$ 
combines various unknown/starred quantities but in the end yields
an observable quantity $\mathcal{T}^*(p,c) =\mathbb{E}[X_p^c] - \mathbb{E}[X_0^c]$
from available embeddings 
($z_p, \psi_c$)
and is therefore amenable to training against empirical risk, as detailed below.
That is, while $\mathcal{D}$ 
does not directly use the true latent representations, 
it aims to recover the true general perturbation effect leveraging the available embeddings. 
We emphasize that we do {\bf not} aim to recover or identify the actual hidden entities 
$f^*, z_p^*, \psi_c^*$ but rather only to predict the output $\mathcal{T}^*(p,c)$.

 To encode a meaningful representation $z_p$ of the perturbation effect, we can invert the above relationship as follows:
\begin{align}\label{eq:true_recovery_of_latents}
    \begin{split}
    z_p &= {h^*_z}^ {-1}(z_p^*)\\
    &= {h^*_z}^ {-1}({f^*}^{-1}(\mathcal{T}^*(p,c)) - \psi_c^*)\\
    &= {h^*_z}^ {-1}({f^*}^{-1}(\mathbb{E}[X_p^c] - \mathbb{E}[X_0^c]) - h_\psi^*(\psi_c))\\
    &= \mathcal{E}_z(\mathbb{E}[X_p^c] - \mathbb{E}[X_0^c],\psi_c ),
    \end{split}
\end{align}

where, in the last line, the function $\mathcal{E}_z$ combines various unknown/starred elements to yield
the proxy latent  $z_p$ (that is used in the decoder $\mathcal{D}$)
from actual observed expression values. 
Notice that $h_z$ is 
necessarily 
invertible, as two perturbations with the same \emph{true} latent representation $z^*$ have the same perturbation effect\footnote{Any true duplicates must behave exactly the same across all cell lines; therefore, in practice we can prune these from the dataset in preprocessing (they do not provide any additional information anyway).}. 
We can define, in an analogous fashion, an encoder $\mathcal{E}_\psi$ for the context.

\medskip

\begin{remark}\label{remark:invariance}
{\it Context invariance.}
Eq. (\ref{eq:true_recovery_of_latents}) implies that
for two different contexts $c,c'$, we have 
$$\mathcal{E}_z(\mathbb{E}[X_p^c] - \mathbb{E}[X_0^c],\psi_c ) = \mathcal{E}_z(\mathbb{E}[X_p^{c'}] - \mathbb{E}[X_0^{c'}],\psi_{c'} ).$$
This is due to the fact that 
given data obtained under perturbation $p$,
the theoretical encoder $\mathcal{E}_z$ yields a latent representation that is specific to perturbation $p$ but invariant across contexts.
\end{remark}

\medskip
{\bf Perturbation effects and empirical risk.}
Putting this all together yields a route towards an applicable architecture to predict perturbation effects and thereby empirical risk. 
First, compute the latent representations $z_p, \psi_c$ from the available data
using the functions $\mathcal{E}_z, \mathcal{E}_\psi$.
 Second, use these representations to recover the perturbation effects $\mathbb{E}[X_p^c] - \mathbb{E}[X_0^c]$. An end-to-end estimate of the actual observed perturbation effects $T^*(c,p)$ is thus:
\begin{align*}
\mathcal{T}(c,p) = \mathcal{D}(&\mathcal{E}_z(\mathbb{E}[X_p^c] - \mathbb{E}[X_0^c],\psi_c ),
\mathcal{E}_\psi(\mathbb{E}[X_p^c] - \mathbb{E}[X_0^c],z_p ))    
\end{align*}

Note that $\mathcal{T}$ is only a function of available data and perturbation and context indices, with all theoretical quantities absorbed into the trainable functions 
$\mathcal{D}, \mathcal{E}_z, \mathcal{E}_\psi$.

This directly implies the following proposition:
\begin{proposition}\label{prop:global_minimizer}
$Y = \mathcal{T}(c,p)$ is a global minimum of the $\ell_2$ loss
$$
\mathcal{L}\left(Y, \mathcal{T}^*(c,p)\right) = \left|\left|Y - \mathcal{T}^*(c,p)\right|\right|_2^2
$$
\end{proposition}
The proof can be found in the appendix in \cref{app:sec:proof_global_minimizer}. 

\section{Architecture}
In this section, we propose a concrete architecture to predict the causal effect 
under a perturbation $p$ in a context $c'$ in which we have never observed the effect of $p$. That is, we aim to predict the \emph{bulk effect} of the perturbation exploiting information on the perturbation and context under the causal manifold assumption in the previous section. 
We focus on the setting in which training data is available as \mbox{(pseudo-)bulk} data $x^c_p = E[X_p^c] - E[X_0^c]$, where $E[\cdot]$ represents the empirical mean.  

Inspired by the theory above, we propose an autoencoder-like architecture. In particular, we learn parametrizations of the encoders $\mathcal{E}_z, \mathcal{E}_\psi$ as well as the decoder $\mathcal{D}$. 
To distinguish empirically learned estimates of these functions from the theoretical 
functions in the previous Section, we write $\hat{\mathcal{E}}_z, \hat{\mathcal{E}}_\psi, \hat{\mathcal{D}}$
for the estimates. 
In our proposed architecture, we set $z_p$ and $\psi_c$ to one-hot encodings of the perturbation and cell line respectively. 

A naive first approach would be to train these modules by a back-propagation pass of each datapoint separately --- as is standard in usual autoencoder architecture.
However, this neglects the context invariance implied by \cref{eq:true_recovery_of_latents} (see \cref{remark:invariance}), namely that: 
\begin{align*}
    \mathcal{E}_z(x^c_p, \psi_c) = z_p = \mathcal{E}_z(x^{c'}_p, \psi_{c'})
\end{align*}
The embedding of the perturbation extracted from the observations by means of the encoder must be the same for each context. This makes sense intuitively, as we aim to extract a representation of the perturbation in general that is \textbf{not} specific to a context. We 
therefore exploit this inductive bias in designing $\hat{\mathcal{E}}_z$, using the following form:
\begin{align*}
    \hat{\mathcal{E}}_z(p) = \bigoplus_{\gamma \in C} \Theta_z(x^{\gamma}_p, \psi_\gamma),
\end{align*}
where $\Theta$ is a neural network and $\bigoplus$ is an aggregation function (in our case we use mean aggregation). 
Similarly, for the encoder embedding the cell line, we have:
\begin{align*}
    \hat{\mathcal{E}}_\psi(c) = \bigoplus_{\pi \in P} \Theta_\psi(x^{c}_\pi, z_\pi)
\end{align*}
Both of these embeddings have the same latent dimension. Thus we can simply combine them using addition, as inspired by \cref{assumption:CMA}. Then, using the neural network $\Gamma$, we can construct the decoder as:
\begin{align*}
    \hat{\mathcal{D}}(\hat{z}_p, \hat{\psi}_c) = \Gamma(\hat{z}_p + \hat{\psi}_c)
\end{align*}
We can now combine the various elements and retrieve a prediction of the perturbation effect as:
\begin{align*}
    \hat{\mathcal{T}}(p,c) = \hat{\mathcal{D}}(\hat{\mathcal{E}}(p), \hat{\mathcal{E}}(c)).
\end{align*}

Note that this quantity is now directly comparable to empirically observed expression. 
Although this method necessitates one whole pass over all contexts in the training dataset for a single perturbation prediction, we can train this model efficiently by computing the forward pass through $\Theta$ in parallel, aggregating ($\bigoplus$) over the batch and again computing the forward pass through $\Gamma$ in parallel.
This architecture can then simply be trained by backpropagating the loss between the prediction and the sample aiming to minimize the $\ell_2$ loss as described in \cref{prop:global_minimizer}.

\begin{table*}[t]
    \centering \large
    \resizebox{\textwidth}{!}{
    \begin{NiceTabular}{l|ccc|ccc|ccc}[colortbl-like]
    \toprule
 & \multicolumn{3}{c}{Replogle} & \multicolumn{3}{c}{Parse} & \multicolumn{3}{c}{Tahoe} \\
Method & Correlation ($\uparrow$) & MSE ($\downarrow$) & Epoch & Correlation ($\uparrow$) & MSE ($\downarrow$) & Epoch & Correlation ($\uparrow$) & MSE ($\downarrow$) & Epoch \\
\midrule
C3TL (Ours) & \cellcolor{best}{0.491} $\pm$ 0.031 & \cellcolor{best}{0.016} $\pm$ 0.005 & 0.25s &\cellcolor{best}{0.670} $\pm$ 0.202 & \cellcolor{best}{0.019} $\pm$ 0.018 & 0.11s& \cellcolor{secondbest}0.777 $\pm$ 0.074 & \cellcolor{best}{0.002} $\pm$ 0.000 & 1.7s\\
State & \cellcolor{secondbest}{0.474 $\pm$ 0.047} & 0.018 $\pm$ \cellcolor{secondbest}0.006 & 10s & \cellcolor{secondbest}0.608 $\pm$ 0.099 & \cellcolor{secondbest} 0.104 $\pm$ 0.028 & 5.4s & \cellcolor{best}{0.778} $\pm$ 0.069 & \cellcolor{secondbest}0.004 $\pm$ 0.003 & 59s\\
CPA                   & 0.008 $\pm$ 0.026 & 0.028 $\pm$ 0.011 & 8.9s & 0.000 $\pm$ 0.000 &  0.589 $\pm$ 0.102 & 3.8s & 0.13 $\pm$ 0.015 & 0.010 $\pm$ 0.003 & 87s \\
Mean Baseline         & 0.370 $\pm$ 0.041 & 62.453 $\pm$ 24.680 & - & 0.627 $\pm$ 0.151 & 4.602 $\pm$ 1.400 & - & 0.725 $\pm$ 0.083 & 1.684 $\pm$ 0.345 & - \\
Closest Cell Baseline & 0.319 $\pm$ 0.059 & 90.141 $\pm$ 30.987 & - & 0.523 $\pm$ 0.147 & 0.926 $\pm$ 0.641 & - & 0.630 $\pm$ 0.067 & 0.682 $\pm$ 0.300 & - \\
\bottomrule
    \end{NiceTabular}}
\caption{\textbf{Performance of the models on perturbation datasets.} The table shows the mean of the Pearson Correlation of the prediction of the individual algorithms with the true target for each gene as well as the Mean Squared Error. Metrics are reported as the mean ($\pm$ standard deviation) over a 5-fold cross validation (4-fold for the Replogle dataset). Best and second best performing methods are colored in blue and gray respectively. In all experiments, metrics are only computed on interventions that have not been seen by the models in the target context.}
    \label{tab:results}
\end{table*} 

\section{Experiments}
\label{sec:experiments}

In this section, we empirically evaluate the proposed method by comparing its predictions against outcomes from biological interventional experiments. We utilize three recent large-scale perturbation datasets \cite{replogle2022mapping, nadig2025transcriptome,zhang2025tahoe,parse10Mpbmcs2026}, covering a diverse range of intervention types and biological contexts. In line with our focus on 
contextual generalization 
in all experiments 
performance
metrics are computed on interventions  that have {\it not} been seen by the models in the target context.

Our evaluation consists of two main parts: first, a benchmarking comparison against state-of-the-art baselines, and second, an analysis of sensitivity and data efficiency using the Tahoe dataset \cite{zhang2025tahoe}. The latter spans a particularly large number of contexts 
and perturbations, thereby enabling 
rigorous empirical study of data efficiency and contextual generalization. 

\medskip
\textbf{Datasets.} The datasets
span several different perturbation modalities: compound perturbations (Tahoe-100, \cite{zhang2025tahoe}), genetic perturbations (Replogle, \cite{replogle2022mapping, nadig2025transcriptome}), and signaling perturbations (Parse, \cite{parse10Mpbmcs2026}). Although these are single-cell datasets containing multiple cellular observations per perturbation, we pseudo-bulk the data for our experiments. Specifically, for each perturbation and cell line combination, we compute the mean gene expression profile across all single cells, aligning with our focus on bulk-level transferability.

The datasets present varying challenges regarding data sparsity and scale. The Tahoe dataset contains 48 contexts (cell lines), each subject to the same 1138 perturbations. The Replogle dataset offers a higher number of perturbations (1677) but is limited to only 4 contexts. Conversely, the Parse dataset includes 24 contexts but features a smaller set of 90 perturbations per context.

\medskip
\textbf{Baselines.} We compare our approach against two established methods: the large-scale foundation model ``State'' \cite{adduri2025predicting} and the autoencoder-based CPA \cite{lotfollahi2023predicting}. Additionally, we include two simple baselines:
\begin{itemize}
    \item \textit{Mean Baseline}: Predicts the average expression of the perturbation $p$ across all training contexts.
    \item \textit{Closest Cell Baseline}: Identifies the training context (cell line) most similar to the test context (based on shared training samples) and predicts the expression observed in that similar context for perturbation $p$.
\end{itemize}

\begin{wrapfigure}{r}{0.6\columnwidth} 
  \begin{center}
    \resizebox{0.6\columnwidth}{!}{\includegraphics[width=\textwidth]{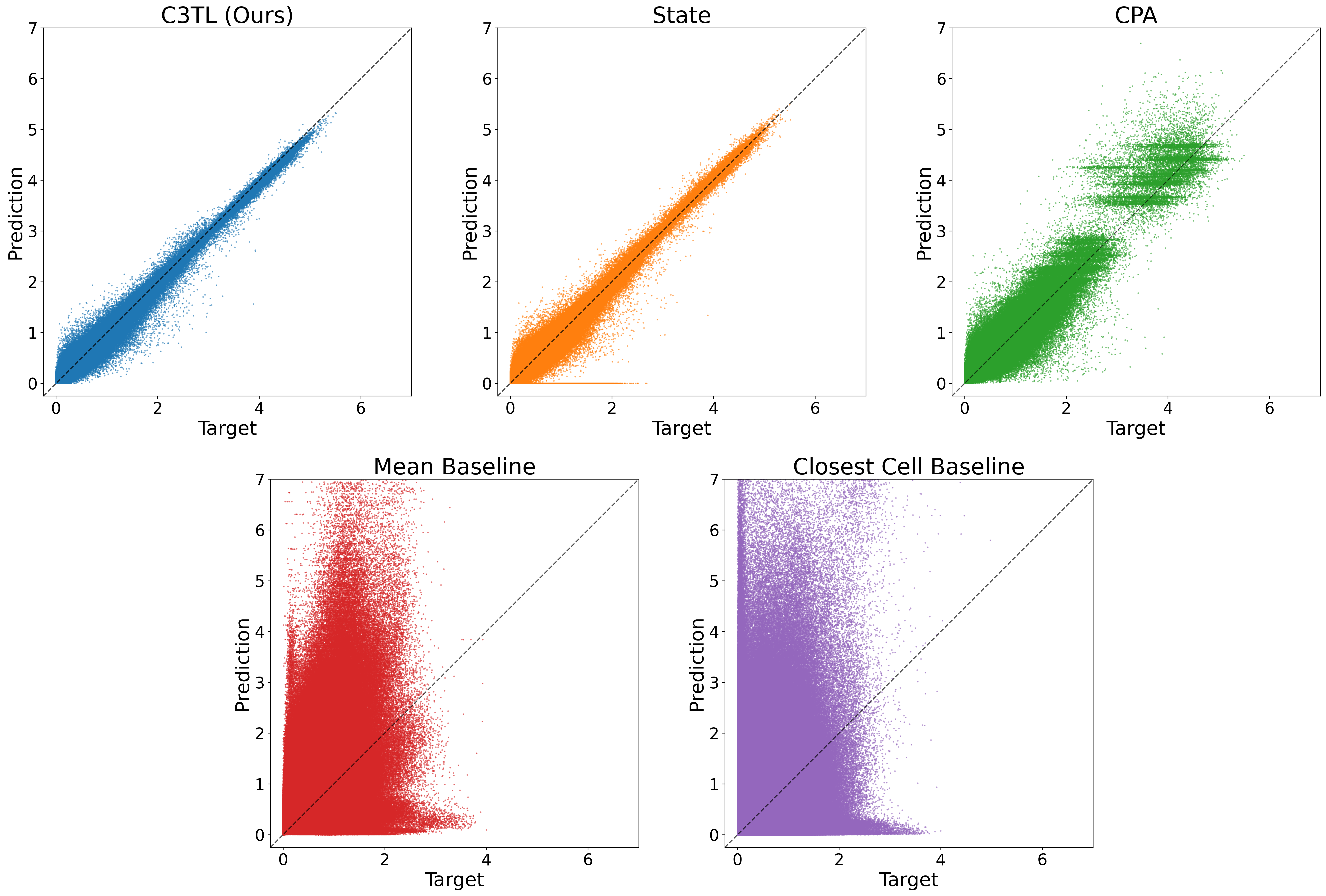}}
\vskip -0.1in
\caption{\textbf{Scatter plot of model predictions.} Comparison of model outputs versus true target values for the Tahoe-100 dataset. The dashed line represents perfect prediction.}
\label{fig:scatter}
\end{center}
\end{wrapfigure}

\medskip
\textbf{Benchmarking Performance.} In our first experiment, we perform cross-validation on the Replogle, Parse, and Tahoe datasets, restricting the gene space to the 2000 most highly variable genes. For the Parse and Tahoe datasets, we randomly select 5 cell lines as held-out test contexts in a leave-one-out fashion. For the Replogle dataset, due to its limited number of cell lines, we employ 4-fold cross-validation. Data from all non-test contexts (cell lines) constitutes the training set. Within the test context, we split the data into 8\% training and 2\% validation (to inform the model about the new context) and reserve 90\% for testing. Note that the specific {\it gene expression} data for perturbations in the test set are never seen by the model during training/validation/testing; only the cell line identity and the perturbation identity are provided in the test case. 
This keeps the focus on causal generalization to interventions that have not yet been seen in the target context.
Due to the significant computational cost of the State model, we performed hyperparameter tuning for all models on the first fold and applied the selected parameters to subsequent folds to ensure a fair comparison. The results are summarized in \cref{tab:results}.
We visually characterize the predictive quality in \cref{fig:scatter}, which plots model predictions against true target values.

\medskip
\textbf{Key observations.} As shown in \cref{tab:results}, our approach is competitive with the current state-of-the-art foundation model State, and achieves superior performance on the Parse dataset. The scatter plot in \cref{fig:scatter} corroborates this, showing that both our method (C$^3$TL) and State closely follow the ideal prediction line. Importantly, our model achieves this accuracy with a considerably more parameter- and compute-efficient architecture. As \cref{tab:results} shows, C$^3$TL achieves an approximately $30\times$ speed up over the other architectures. Indeed, C$^3$TL can even be run without a GPU in reasonable time. Additionally, State uses approx. $19.8$GB , CPA uses approx. $0.5$GB and C$^3$TL uses approx. $2.1$GB GPU memory, making it about $9\times$ more memory efficient than State. 

\medskip
\textbf{Data efficiency and performance in limited data regimes.} We further investigate model performance in data-constrained scenarios. 
Data-constrained scenarios
are highly relevant in the application of deep perturbation models in real biomedical use-cases. This is due to the fact that in a wide range of specific disease-related and biological settings, only limited data can be generated due to cost and other constraints. In that sense, comprehensive perturbation atlases like Tahoe are not representative 
of typical biological and medical use-cases. 

\begin{figure*}[t]
\vskip 0.2in
\begin{center}
\resizebox{\textwidth}{!}{\includegraphics[width=\textwidth]{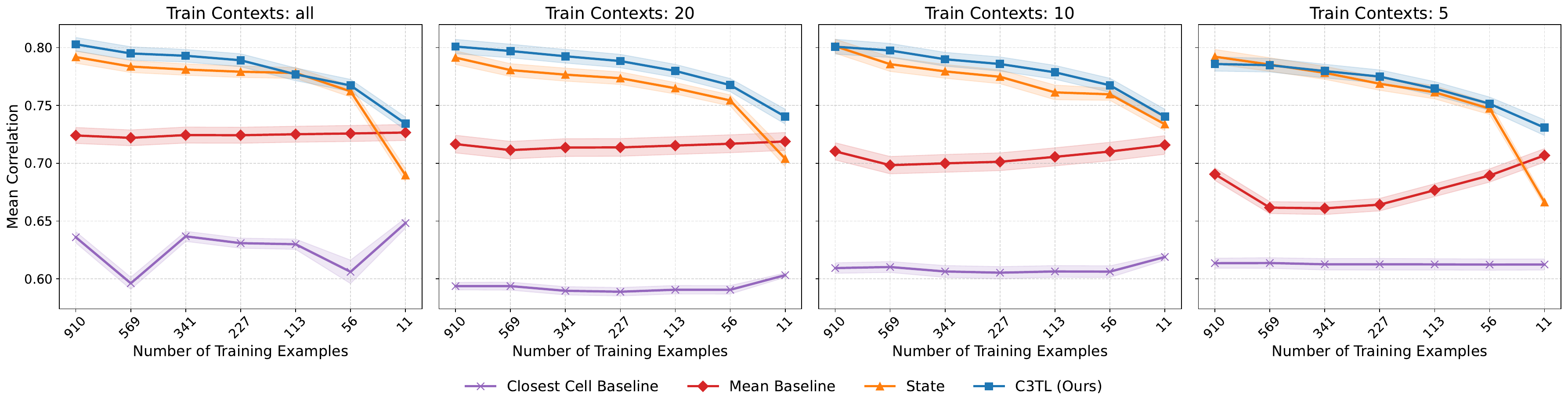}}
\vskip -0.15in
\caption{\textbf{Sensitivity to data scarcity.} Performance evaluation under varying training data availability. Panels (left to right) display regimes with 43, 20, 10, and 5 randomly sampled training contexts. The x-axis represents the number of 
interventions in the test context available for adaptation (implying the complement was used for testing). This ticks correspond to (from left to right) $80\%$, $50\%$, $30\%$, $20\%$, $10\%$, $5\%$, $1\%$ of the data available in the test contexts in total. The y-axis displays the Pearson Correlation between prediction and target ($\pm$ standard deviation across 5 test contexts).
In all experiments, metrics are only computed on interventions that have not been seen by the models in the target context.}
\label{fig:sensitivity}
\end{center}
\end{figure*}

To 
better understand performance in the data-constrained setting,
we systematically varied both the number of available training contexts and the
number of interventions/data 
available for adaptation
in target contexts. 
For these experiments we focus on the Tahoe dataset, whose relatively large number of contexts and perturbations makes
it ideal for systematic evaluation of these aspects.
We randomly selected 5 contexts (cell lines) as held-out test contexts. From the remaining 43 contexts, we randomly created training sets of size 43, 20, 10, and 5. To ensure a more controlled comparison, these sets were nested: contexts in smaller training sets were included in larger ones. Within the test contexts, we varied the adaptation data (i.e. the interventions available in the target context)
by splitting samples into $X\%/(100-X)\%$ adaptation/test splits. The adaptation portion was further divided 75\%/25\% for training and validation. We also enforced a nested structure on the test sets, ensuring that perturbations in a smaller test set were included in larger test sets. 
As in all our experiments, performance
metrics are only computed on interventions  that have not been seen by the models in the target context.
Results appear in \cref{fig:sensitivity}. CPA is excluded for better legibility.

\medskip
\textbf{Key observations.} \cref{fig:sensitivity} demonstrates that both State and our approach 
C$^3$TL
exhibit robustness to data scarcity. However, C$^3$TL shows a distinct advantage in the extremely scarce regime (5 training contexts, 99\% testing data; this means adaptation has 
to be done using only 1\% of interventions in the target context). This demonstrates that 
within the Tahoe atlas
a very small number of observations (10-100 perturbations) 
in a target context
is sufficient for our model to generate reliable predictions, highlighting its practical utility for data-limited biomedical  applications.

\section{Discussion}

We proposed a framework, C$^3$TL, for the modelling of quantitative causal effects in large-scale biology, that leverages a notion of causal manifolds to allow generalization to novel contexts. We showed that the C$^3$TL architecture is compute- and data-efficient and competitive with a large-scale foundation model, with particuarly strong performance in the limited data regimes typical of many biomedical applications. There are several key open questions, limitations and directions for future work that we highlight below.

Our approach provides a scalable but relatively efficient  way to model quantitative causal effects in biology and could be used in hybrid computational-experimental workflows to efficiently investigate perturbation effects in novel contexts. However, doing so would involve further developments within an active learning or reinforcement learning framework, in which C$^3$TL provides the core model and the action space spans all possible perturbations. 

Our theory focused on information flow and manifold assumptions concerning how perturbation and context relate to each other in an idealized latent space. 
A rich literature in causal theory focuses on formalization via structural causal models (SCMs), including at the ML interface. A detailed understanding of our framework from an SCM perspective would be interesting in its own right and might suggest new avenues for architectural improvements.

More generally, our theory is limited in that the focus was on existence of required functions rather than 
learning theory (e.g. formal finite-sample results concerning data-efficiency). We saw empirically that C$^3$TL is both effective and, in line with its model structure, highly data efficient. Building on the CMA assumption, it would be useful to understand data efficiency from a theoretical point of view, relating model performance to the number and type of the various inputs.

\clearpage

{\bf Acknowledgements.} An LLM was used to assist in generating code to produce Figure 1.

\bibliographystyle{unsrt}  
\bibliography{example_paper}  

@article{lotfollahi2023predicting,
  title={Predicting cellular responses to complex perturbations in high-throughput screens},
  author={Lotfollahi, Mohammad and Klimovskaia, Anna and Orr, Carlo and others},
  journal={Molecular Systems Biology},
  volume={19},
  number={6},
  pages={e11517},
  year={2023},
  publisher={EMBO Press}
}

@article{adduri2025predicting,
  title={{Predicting cellular responses to perturbation across diverse contexts with State}},
  author={Adduri, N. and others},
  journal={bioRxiv},
  year={2025},
  publisher={Cold Spring Harbor Laboratory}
}

@article{zhang2025tahoe,
  title={{Tahoe-100m: A giga-scale single-cell perturbation atlas for context-dependent gene function and cellular modeling}},
  author={Zhang, Jesse and Ubas, Airol A and de Borja, Richard and Svensson, Valentine and Thomas, Nicole and Thakar, Neha and Lai, Ian and Winters, Aidan and Khan, Umair and Jones, Matthew G and others},
  journal={BioRxiv},
  pages={2025--02},
  year={2025},
  publisher={Cold Spring Harbor Laboratory}
}

@article{replogle2022mapping,
  title={{Mapping information-rich genotype-phenotype landscapes with genome-scale Perturb-seq}},
  author={Replogle, Joseph M and Saunders, Reuben A and Pogson, Angela N and Hussmann, Jeffrey A and Lenail, Alexander and Guna, Alina and Mascibroda, Lauren and Wagner, Eric J and Adelman, Karen and Lithwick-Yanai, Gila and others},
  journal={Cell},
  volume={185},
  number={14},
  pages={2559--2575},
  year={2022},
  publisher={Elsevier}
}

@misc{parse10Mpbmcs2026,
  title        = {{10 Million Human PBMCs in a Single Experiment}},
  howpublished = {\url{https://www.parsebiosciences.com/datasets/10-million-human-pbmcs-in-a-single-experiment/}},
  author       = {{Parse Biosciences}},
  year         = {2026},
  note         = {Accessed: 2026-01-07},
}

@article{lotfollahi2019scgen,
  title={{scGen predicts single-cell perturbation responses}},
  author={Lotfollahi, Mohammad and Wolf, F Alexander and Theis, Fabian J},
  journal={Nature Methods},
  volume={16},
  number={8},
  pages={715--721},
  year={2019},
  publisher={Nature Publishing Group}
}

@article{roohani2023gears,
  title={{Predicting transcriptional outcomes of novel multigene perturbations with GEARS}},
  author={Roohani, Yusuf and Huang, Kexin and Leskovec, Jure},
  journal={Nature Biotechnology},
  year={2023},
  publisher={Nature Publishing Group}
}

@article{hetzel2022predicting,
  title={{Predicting the effect of chemical perturbations with ChemCPA}},
  author={Hetzel, Leon and B{\"o}hm, Simon and Kilbertus, Niki and G{\"u}nnemann, Stephan and Theis, Fabian J and Lotfollahi, Mohammad},
  journal={Bioinformatics},
  year={2022}
}

@article{hill2019causal,
  title={Causal learning via manifold regularization},
  author={Hill, Steven M and Oates, Chris J and Blythe, Duncan A and Mukherjee, Sach},
  journal={Journal of Machine Learning Research},
  volume={20},
  number={127},
  pages={1--32},
  year={2019}
}

@article{subramanian2017l1000,
  title={{A next generation connectivity map: L1000 platform and the first 1,000,000 profiles}},
  author={Subramanian, Aravind and Narayan, Rajiv and Corsello, Steven M and others.},
  journal={Cell},
  volume={171},
  number={6},
  pages={1437--1452},
  year={2017},
  publisher={Elsevier}
}

@article{cui2024scgpt,
  title={{scGPT: toward building a foundation model for single-cell multi-omics using generative AI}},
  author={Cui, Haotian and Wang, Chloe and Maan, Hassaan and Pang, Kuan and Luo, Fengning and Duan, Nan and Wang, Bo},
  journal={Nature Methods},
  year={2024}
}

@article{iwata2019predicting,
  title={Predicting drug-induced transcriptome responses of a wide range of human cell lines by a novel tensor-train decomposition algorithm},
  author={Iwata, Michio and Yuan, Lu and Zhao, Qibin and Tabei, Yasuo and Berenger, Fran{\c{c}}ois and Sawada, Ryusuke and Akiyama, Yutaka and Hamada, Michiaki},
  journal={Bioinformatics},
  volume={35},
  number={14},
  pages={i191--i199},
  year={2019},
  publisher={Oxford University Press}
}

@inproceedings{hodos2018cell,
  title={Cell-specific prediction and application of drug-induced gene expression profiles},
  author={Hodos, Rachel and Zhang, Ping and Lee, Hsuan-Chao and Duan, Qiaonan and Wang, Zichen and Clark, Neil R and Ma'ayan, Avi and Kidd, Brian A and Dudley, Joel T},
  booktitle={Pacific Symposium on Biocomputing},
  volume={23},
  pages={32--43},
  year={2018},
  organization={World Scientific}
}

@inproceedings{pfister2022identifiability,
  title={Identifiability of sparse causal effects using instrumental variables},
  author={Pfister, Niklas and Peters, Jonas},
  booktitle={Uncertainty in Artificial Intelligence},
  pages={1613--1622},
  year={2022},
  organization={PMLR}
}

@article{lagemann2023deep,
  title={Deep learning of causal structures in high dimensions under data limitations},
  author={Lagemann, Kai and Lagemann, Christian and Taschler, Bernd and Mukherjee, Sach},
  journal={Nature Machine Intelligence},
  volume={5},
  number={11},
  pages={1306--1316},
  year={2023},
  publisher={Nature Publishing Group UK London}
}

@article{ke2022learning,
  title={Learning to induce causal structure},
  author={Ke, Nan Rosemary and Chiappa, Silvia and Wang, Jane and Goyal, Anirudh and Bornschein, Jorg and Rey, Melanie and Weber, Theophane and Botvinic, Matthew and Mozer, Michael and Rezende, Danilo Jimenez},
  journal={arXiv preprint arXiv:2204.04875},
  year={2022}
}

@inproceedings{lippe2022,
  title={Efficient neural causal discovery without acyclicity constraints.},
  author={Lippe, Phillip and Cohen, Taco and Gavves, Efstratios},
  booktitle={International Conference on Learning Representations},
  year={2022},
}

@article{lopez2022large,
  title={Large-scale differentiable causal discovery of factor graphs},
  author={Lopez, Romain and H{\"u}tter, Jan-Christian and Pritchard, Jonathan and Regev, Aviv},
  journal={Advances in Neural Information Processing Systems},
  volume={35},
  pages={19290--19303},
  year={2022}
}

@article{heinze2018causal,
  title={Causal structure learning},
  author={Heinze-Deml, Christina and Maathuis, Marloes H and Meinshausen, Nicolai},
  journal={Annual Review of Statistics and Its Application},
  volume={5},
  number={1},
  pages={371--391},
  year={2018},
  publisher={Annual Reviews}
}

@article{dixit2016perturb,
  title={{Perturb-Seq: dissecting molecular circuits with scalable single-cell RNA profiling of pooled genetic screens}},
  author={Dixit, Atray and Parnas, Oren and Li, Biyu and Chen, Jenny and Fulco, Charles P and Jerby-Arnon, Livnat and Marjanovic, Nemanja D and Dionne, Danielle and Burks, Tyler and Raychowdhury, Raktima and others},
  journal={Cell},
  volume={167},
  number={7},
  pages={1853--1866},
  year={2016},
  publisher={Elsevier}
}

@article{nadig2025transcriptome,
  title={Transcriptome-wide analysis of differential expression in perturbation atlases},
  author={Nadig, Ajay and Replogle, Joseph M and Pogson, Angela N and Murthy, Mukundh and McCarroll, Steven A and Weissman, Jonathan S and Robinson, Elise B and O’Connor, Luke J},
  journal={Nature Genetics},
  pages={1--10},
  year={2025},
  publisher={Nature Publishing Group US New York}
}


\newpage
\appendix
\onecolumn
\section{Dataset Details}
\begin{table*}[h]
\caption{Overview of datasets used for evaluation. The datasets cover different perturbation modalities including compound, genetic, and signaling perturbations.}
\label{tab:datasets}
\vskip 0.15in
\begin{center}
\begin{small}
\begin{sc}
\begin{tabular}{lccc}
\toprule
 & Tahoe-100 & Replogle & Parse \\
\midrule
Type & Compound & Genetic & Signaling \\
\# Contexts & 48 & 4 & 24 \\
\# Pert. / Context & 1138 & 1677 & 90 \\
Reference & \cite{zhang2025tahoe} & \cite{replogle2022mapping, nadig2025transcriptome} & \cite{parse10Mpbmcs2026} \\
\bottomrule
\end{tabular}
\end{sc}
\end{small}
\end{center}
\vskip -0.1in
\end{table*}

\section{Proofs}
\subsection{Proof of \cref{prop:global_minimizer}}\label{app:sec:proof_global_minimizer}
To prove that 
$$ \left|\left|Y - \mathcal{T}^*(c,p)\right|\right|_2^2 $$
is the global minimizer of the $\ell_2$ norm for $Y = \mathcal{T}^*(c,p)$, we show that 
$$ \left|\left|\mathcal{T}(c,p) - \mathcal{T}^*(c,p)\right|\right|_2^2 = 0 .$$
This is the case if and only if $\mathcal{T}(c,p) = \mathcal{T}^*(c,p)$. Which we show by direct computation: 
\begin{align*}
    \mathcal{T}(c,p) =\mathcal{D}(\mathcal{E}_z(\mathbb{E}[X_p^c] - \mathbb{E}[X_0^c],\psi_c ),  \mathcal{E}_\psi(\mathbb{E}[X_p^c] - \mathbb{E}[X_0^c],z_p ))  
\end{align*}
This holds by definition of $\mathcal{T}(c,p)$. 
For $\mathcal{E}_z(\mathbb{E}[X_p^c] - \mathbb{E}[X_0^c],\psi_c )$, we have:
\begin{align*}
\mathcal{E}_z(\mathbb{E}[X_p^c] - \mathbb{E}[X_0^c],\psi_c )
    & \overset{\text{def.}}{=} {h^*_z}^ {-1}({f^*}^{-1}(\mathbb{E}[X_p^c] - \mathbb{E}[X_0^c]) - h_\psi^*(\psi_c))\\
    & \overset{(\ref{eq:true_pert_effect_equals_diff_of_expectation})}{=} {h^*_z}^ {-1}({f^*}^{-1}(\mathcal{T}^*(p,c)) - \psi_c^*)\\
    & \overset{\ref{assumption:CMA}}{=} {h^*_z}^ {-1}(z_p^*)\\
    & \overset{(\ref{eq:def_correction})}{=} z_p 
\end{align*}
The proof for $\mathcal{E}_\psi(\mathbb{E}[X_p^c] - \mathbb{E}[X_0^c],z_p ) = \psi_c$ is analogous.
Thus, 
\begin{align*}
    \mathcal{T}(c,p) &=\mathcal{D}(z_p, \psi_c) \\
    & \overset{\text{def.}}{=}  f^*(h_z^*(z_p) + h_\psi^*(\psi_c)) \\
    & \overset{(\ref{eq:def_correction})}{=} f^*(z_p^* + \psi_c^*) \\
    & \overset{\ref{assumption:CMA}}{=} \mathcal{T}^*(p,c) 
\end{align*} \qed

\end{document}